\documentclass[acmsmall,screen]{acmart}
\usepackage{amsfonts}
\usepackage{threeparttable}
\usepackage{multirow}
\usepackage{multicol}
\usepackage{makecell}
\usepackage{algpseudocode}
\usepackage{booktabs}
\usepackage{graphicx}
\usepackage{subfigure}
\newcommand{\tabincell}[2]{\begin{tabular}{@{}#1@{}}#2\end{tabular}}
\usepackage{algorithm}
\usepackage{amsmath}
\usepackage{hyperref}
\usepackage{bbm}

\algnewcommand\algorithmicforeach{\textbf{for each}}
\algdef{S}[FOR]{ForEach}[1]{\algorithmicforeach\ #1\ \algorithmicdo}

\AtBeginDocument{%
  \providecommand\BibTeX{{%
    \normalfont B\kern-0.5em{\scshape i\kern-0.25em b}\kern-0.8em\TeX}}}

\setcopyright{acmlicensed}
\copyrightyear{2024}
\acmYear{2024}
\acmDOI{XXXXXXX.XXXXXXX}

\acmJournal{TOIS}
\acmVolume{1}
\acmNumber{1}
\acmArticle{1}
\acmMonth{12}

\begin{document}

\title{A Debiased Nearest Neighbors Framework for Multi-Label Text Classification}
%
\author{Zifeng Cheng}
\email{chengzf@smail.nju.edu.cn}
\orcid{0000-0002-8486-2614}
\author{Zhiwei Jiang}
\authornote{Corresponding Author.}
\email{jzw@nju.edu.cn}
\orcid{0000-0001-5243-4992}
\author{Yafeng Yin}
\email{yafeng@nju.edu.cn}
\orcid{0000-0002-9497-6244}
\author{Zhaoling Chen}
\email{zhaolingchen@smail.nju.edu.cn}
\orcid{0009-0000-0041-5255}
\author{Cong Wang}
\email{cw@smail.nju.edu.cn}
\orcid{0000-0003-0916-7803}
\author{Shiping Ge}
\email{shipingge@smail.nju.edu.cn}
\orcid{0000-0001-9198-5324}
\affiliation{%
  \thanks{This work is supported by the National Natural Science Foundation of China under Grants Nos. 61972192, 62172208, 61906085.
This work is partially supported by Collaborative Innovation Center of Novel Software Technology and Industrialization. Zhiwei Jiang is the corresponding author.}
  \institution{State Key Laboratory for Novel Software Technology, Nanjing University}
  \streetaddress{163 Xianlin Ave}
  \city{Nanjing}
  \state{Jiangsu}
  \country{China}
  \postcode{210023}
}
\author{Qiguo Huang}
\email{huangqiguo2003@126.com}
\orcid{0000-0001-7912-7175}
\affiliation{%
  \institution{Nanjing Tuqin Artificial Intelligence Research Institute Co., LTD}
  \city{Nanjing}
  \state{Jiangsu}
  \country{China}
  \postcode{210023}
}
\author{Qing Gu}
\email{guq@nju.edu.cn}
\orcid{0000-0002-1112-790X}
\affiliation{%
  \institution{State Key Laboratory for Novel Software Technology, Nanjing University}
  \streetaddress{163 Xianlin Ave}
  \city{Nanjing}
  \state{Jiangsu}
  \country{China}
  \postcode{210023}
}

\renewcommand{\shortauthors}{Cheng, et al.}

\begin{abstract}
Multi-Label Text Classification (MLTC) is a practical yet challenging task that involves assigning multiple non-exclusive labels to each document. 
Previous studies primarily focus on capturing label correlations to assist label prediction by introducing special labeling schemes, designing specific model structures, or adding auxiliary tasks.
Recently, the $k$ Nearest Neighbor ($k$NN) framework has shown promise by retrieving labeled samples as references to mine label co-occurrence information in the embedding space. 
However, two critical biases, namely embedding alignment bias and confidence estimation bias, are often overlooked, adversely affecting prediction performance.
In this paper, we introduce a DEbiased Nearest Neighbors (DENN) framework for MLTC, specifically designed to mitigate these biases. 
To address embedding alignment bias, we propose a debiased contrastive learning strategy, enhancing neighbor consistency on label co-occurrence. 
For confidence estimation bias, we present a debiased confidence estimation strategy, improving the adaptive combination of predictions from $k$NN and inductive binary classifications. 
Extensive experiments conducted on four public benchmark datasets (i.e., AAPD, RCV1-V2, Amazon-531, and EUR-LEX57K) showcase the effectiveness of our proposed method.
Besides, our method does not introduce any extra parameters.
\end{abstract}





\maketitle

\section{Introduction}

Multi-Label Text Classification (MLTC) is a basic and long-standing task that aims to assign multiple non-exclusive labels to each document from a predefined label set.
For example, the first text is about document representation and latent semantic indexing, and the second text is about energy-efficient power control in Table~\ref{example}.
Thus, MLTC needs to classify the first text as ``cs.CL'' and ``cs.IR'', and the second text as ``cs.IT'' and ``math.IT''.
It is a practical task beneficial for users to retrieve useful documents.
A simple and straightforward solution for MLTC is to conduct independent binary classification for each label, whereas the correlation between labels is ignored.
Taking label correlations into consideration to further improve performance poses a challenge.

Existing methods primarily focus on capturing label correlations to assist label prediction by introducing special labeling schemes, designing specific model structures, or adding auxiliary tasks. 
Regarding labeling schemes, researchers attempt to transform the MLTC task into sequence-to-sequence~\cite{nam2017maximizing,DBLP:conf/coling/YangSLMWW18,tsai2020order,DBLP:conf/emnlp/LinSYM018}, sequence-to-set~\cite{DBLP:conf/acl/YangLMLS19}, and iterative reasoning~\cite{wang-etal-2021-meta} problems to capture label dependency along with the labeling process. 
Regarding model structures, various neural components are designed to explicitly capture label correlation, such as label attention~\cite{DBLP:conf/acl/HenaoLCSWWZZ18,DBLP:conf/aaai/DuCF0GN19}, graph neural network~\cite{DBLP:conf/acl/MaYZH21}, and global embedding~\cite{DBLP:conf/ijcai/ZhangZYLCZ21,DBLP:conf/eacl/AlhuzaliA21}. 
Regarding auxiliary tasks, many relevant tasks are also introduced to implicitly incorporate label correlation into document representation, such as pairwise and conditional label co-occurrence prediction~\cite{DBLP:conf/acl/ZhangZYLC21} and contrastive learning~\cite{DBLP:conf/acl/LinQWZY23}.

\begin{table}[t]
\caption{Two examples for multi-label text classification.} \label{example}
\begin{tabular}{p{10cm}|p{3cm}}
\toprule
\textbf{Text} & \textbf{Labels} \\
\midrule
We consider the problem of creating document representations in which inter-document similarity measurements correspond to semantic similarity. We first present a novel subspace-based framework for formalizing this task. Using this framework, we derive a new analysis of Latent Semantic Indexing (LSI), showing a precise relationship between its performance and the uniformity of the underlying distribution of documents over topics. ... & Computer Science. Computation and Language (cs.CL), Computer Science. Information Retrieval (cs.IR)\\
\midrule
A unified approach to energy-efficient power control, applicable to a large family of receivers including the matched filter, the decorrelator, the (linear) minimum-mean-square-error detector (MMSE), and the individually and jointly optimal multiuser detectors, has recently been proposed for code-division-multiple-access (CDMA) networks. This unified power control (UPC) algorithm exploits the linear relationship that has been shown to exist between the transmit power and the output signal-to-interference-plus-noise ratio (SIR) in large systems. ... & Computer Science. Information Theory (cs.IT), Mathematics. Information Theory (math.IT)\\
\bottomrule
\end{tabular}
\end{table}

Recently, $k$ Nearest Neighbor ($k$NN) framework~\cite{DBLP:conf/acl/SuWD22} has shown promise by directly retrieving labeled samples as references to mine label co-occurrence information in the embedding space. 
In this framework, the labels of each training sample can be viewed as a special case of label co-occurrence.
Then, samples located near the target in the embedding space can provide valuable information of label co-occurrence, resulting in a personalized $k$NN prediction for the target.
This prediction carries label correlation and can be further combined with the prediction from inductive binary classifications to form the final prediction.

\begin{figure}[t]
\centering
\includegraphics[width=0.8\columnwidth]{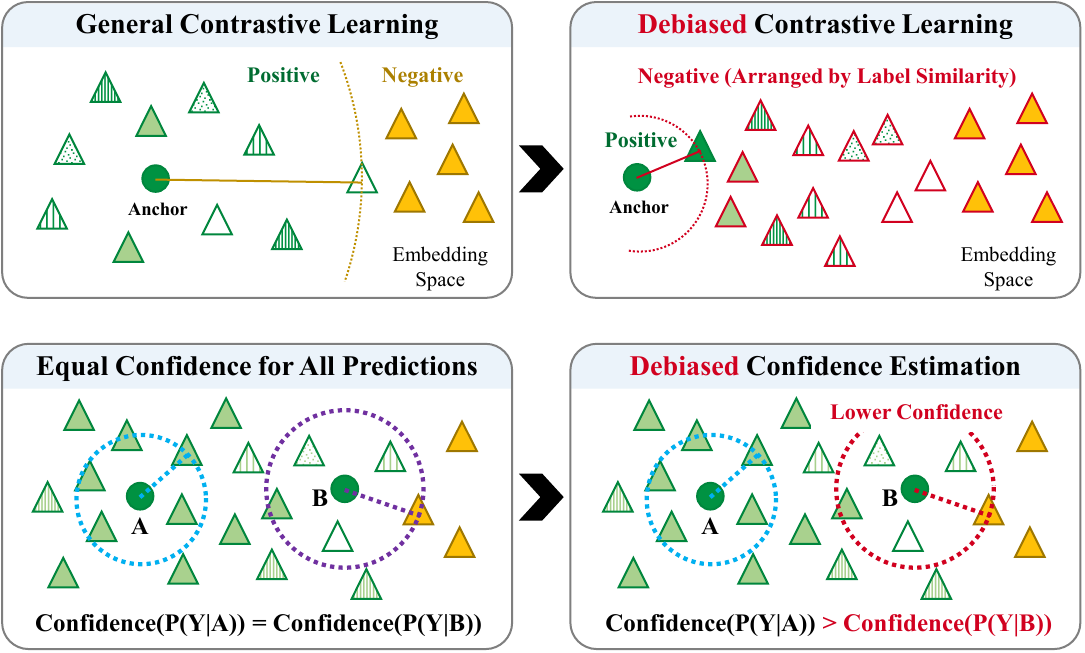}
\caption{Illustration of two bias problems and corresponding solutions. The degree of greenness of each sample is proportional to the label similarity to the anchor. The yellow samples indicate that they do not have the same label as the anchor.} \label{fig:moti}
\end{figure}

This $k$NN framework provides a convenient manner to capture personalized label co-occurrence information through neighbor retrieval, avoiding the complex induction of general label correlation rules in inductive methods.
However, two critical biases are often overlooked, adversely affecting prediction performance.
Firstly, the generally adopted contrastive learning is unable to effectively reflect the similarity on label co-occurrence using embedding similarity, which we denote as \emph{embedding alignment bias}. 
As shown in the upper-left part of Figure ~\ref{fig:moti}, roughly treating all samples with overlapped labels as positive samples, regardless of the degree of overlap, can compromise the embedding space's ability to maintain neighbor consistency on label co-occurrence.
By addressing this bias, $k$NN retrieval can obtain more accurate label co-occurrence information for prediction.
Secondly, it is assumed by default that all $k$NN predictions have the same confidence, which we denote as \emph{confidence estimation bias}.
As shown in the lower part of Figure ~\ref{fig:moti}, retrieved neighbors with more concentrated distribution and more consistent labels often provide more credible label co-occurrence information.
By addressing this bias, the confidence of $k$NN prediction in the final prediction combination can be adaptively estimated for each target sample specifically.

In this paper, we introduce a DEbiased Nearest Neighbors (DENN) framework for MLTC, specifically designed to mitigate these biases. 
Specifically, to address embedding alignment bias, we propose a debiased contrastive learning strategy, enhancing neighbor consistency on label co-occurrence. 
As shown in the upper-right part of Figure ~\ref{fig:moti}, we set the augmented representation of the anchor as positive while all other samples are negatives and arranged based on their label similarity to the anchor.
To address confidence estimation bias, we present a debiased confidence estimation strategy, providing more accurate and adaptive confidence for combining predictions.
To quantify the confidence of $k$NN prediction in the final prediction combination, we comprehensively consider the distribution concentration and label consistency of the retrieved neighbors, along with probability distribution in both predictions.
Based on these two strategies, for each target sample, we can effectively retrieve its personalized label co-occurrence information in the embedding space and dynamically estimate the confidence of such information, resulting in an adaptive label prediction.

The main contributions of this paper can be summarized as follows:
\begin{itemize}
\item We propose a DENN framework for MLTC, using debiased contrastive learning to adjust the biased embedding space for better $k$NN retrieval and using debiased confidence estimation to estimate the confidence of $k$NN.
\item We propose a debiased contrastive learning strategy for MLTC to solve \emph{false positives} problem while controlling the distribution of negatives.
\item We propose a debiased confidence estimation strategy to avoid using constant confidence for all $k$NN retrieval.
\item We conduct experiments on the benchmark datasets (i.e., AAPD, RCV1-V2, Amazon-531, and EUR-LEX57K). The experimental results demonstrate that our proposed method outperforms previous methods and achieves state-of-the-art performance. Besides, our method does not introduce any extra parameters.
\end{itemize}

\section{Related Work}
In this section, we introduce the following three research topics relevant to our work: multi-label text classification, nearest neighbor based retrieval-augmented method, and contrastive learning.

\subsection{Multi-Label Text Classification}
MLTC aims to assign related labels given a text and has wide applications in many areas \cite{DBLP:conf/kdd/ChangJYTZZKHSIS21,DBLP:conf/sigir/JiangCZHY22,DBLP:conf/www/GeJCWYG23}.
Binary Relevance (BR)~\cite{DBLP:journals/pr/BoutellLSB04}, ML-KNN~\cite{DBLP:journals/pr/ZhangZ07} and Classifier Chains (CC)~\cite{DBLP:journals/ml/ReadPHF11} are three classic early multi-label works.
BR~\cite{DBLP:journals/pr/BoutellLSB04} decomposes the MLTC task into multiple independent binary classification problem without considering the correlations between labels and trains one binary classifier (linear SVM) for each label.
ML-KNN~\cite{DBLP:journals/pr/ZhangZ07} is the multi-label version of $k$NN algorithm.
ML-KNN first obtains statistical information gained from the label sets of these neighboring instances in the training set, i.e., the number of neighboring instances belonging to each possible class, and then uses the maximum a posteriori (MAP) principle to determine the label set for the test text.
CC~\cite{DBLP:journals/ml/ReadPHF11} transforms the multi-label learning problem into a chain of binary classification problems, where subsequent binary classifiers in the chain are built upon the predictions of preceding ones.
Recently, with the development of neural networks, many neural network models have been proposed to solve the MLTC task.
Most works in MLTC can be roughly divided into four major categories to capture label correlations: designing specific model structures~\cite{DBLP:conf/acl/HenaoLCSWWZZ18,DBLP:conf/aaai/DuCF0GN19,DBLP:conf/emnlp/XiaoHCJ19,DBLP:conf/ijcai/ZhangZYLCZ21,DBLP:conf/acl/MaYZH21}, introducing special labeling schemes~\cite{nam2017maximizing,DBLP:conf/emnlp/LinSYM018,DBLP:conf/coling/YangSLMWW18,DBLP:conf/acl/YangLMLS19,tsai2020order}, adding auxiliary tasks~\cite{DBLP:conf/acl/ZhangZYLC21,DBLP:conf/acl/LinQWZY23}, retrieving labeled training samples~\cite{DBLP:conf/acl/SuWD22}.

The first type of approach designs specific model structures to mix label information for capturing label correlations.
Specifically, LEAM~\cite{DBLP:conf/acl/HenaoLCSWWZZ18}, EXAM~\cite{DBLP:conf/aaai/DuCF0GN19}, and LSAN~\cite{DBLP:conf/emnlp/XiaoHCJ19} all use label embedding to model fine-grained matching signals between words and labels.
LDGN~\cite{DBLP:conf/acl/MaYZH21} uses the graph neural network to capture the semantic interactions between labels.
CORE~\cite{DBLP:conf/ijcai/ZhangZYLCZ21} and SpanEmo \cite{DBLP:conf/eacl/AlhuzaliA21} both propose a basic global embedding strategy that represents context and all labels in the same latent space and feed them into BERT to capture correlations between labels.

The second type of approach introduces special labeling schemes to capture label correlations, including sequence-to-sequence learning~\cite{nam2017maximizing,DBLP:conf/coling/YangSLMWW18,DBLP:conf/emnlp/LinSYM018,tsai2020order}, sequence-to-set learning~\cite{DBLP:conf/acl/YangLMLS19}, and iterative reasoning~\cite{DBLP:journals/ipm/WangRSQD21}.
Nam et al.~\cite{nam2017maximizing} first propose a sequence-to-sequence learning framework to solve the MLTC task.
SGM~\cite{DBLP:conf/coling/YangSLMWW18} further uses attention mechanism and global embedding to improve the performance.
Lin et al.~\cite{DBLP:conf/emnlp/LinSYM018} further propose a hybrid attention mechanism to capture information at different levels (i.e., word-level, phrase-level, and sentence-level) due to the attention mechanism does not play a significant role in MLTC.
Tsai and Lee~\cite{tsai2020order} use the reinforcement learning algorithm based on optimal completion distillation to mitigate both exposure bias and label order in the sequence-to-sequence learning framework.
Yang et al.~\cite{DBLP:conf/acl/YangLMLS19} argue that the output labels are essentially an unordered set rather than an ordered sequence and propose a sequence-to-set model using reinforcement learning to train.
ML-R~\cite{DBLP:journals/ipm/WangRSQD21} proposes a novel iterative reasoning mechanism that takes account of the text and the label predictions from the previous reasoning step.

The third type of approach adds auxiliary tasks to capture label correlations.
LACO~\cite{DBLP:conf/acl/ZhangZYLC21} further explores two feasible label correlations tasks (i.e., predict whether a label appears given a specific positive label and a set of positive labels) based on CORE framework.
Lin et al. \cite{DBLP:conf/acl/LinQWZY23} further explore five different contrastive losses and found that strictly contrastive loss and Jaccard similarity contrastive loss perform better.

The fourth group of methods retrieves labeled training samples to capture label correlations.
Su et al.~\cite{DBLP:conf/acl/SuWD22} use $k$NN framework which retrieves labels of $k$NN and interpolates the classifier output with labels of $k$NN during the inference phase.
To improve the quality of retrieved neighbors, they use contrastive learning to train the model.
Specifically, they treat samples with one identical label as positives and design contrastive learning with a dynamic coefficient for each sample pair.
Different from Su et al.~\cite{DBLP:conf/acl/SuWD22}, we propose a debiased contrastive learning for MLTC to avoid \emph{false positives} and make the embedding similarities of negatives correlate with the label similarity, and adaptively combine the results of the classifier and $k$NN.

In addition, some works focus on hierarchical multi-label text classification (i.e., labels organized in a hierarchical taxonomy)~\cite{DBLP:conf/acl/BanerjeeAPT19,DBLP:conf/acl/ZhouMLXDZXL20}, extreme multi-label text classification (i.e., the number of labels can reach hundreds of thousands or millions)~\cite{you2019attentionxml,DBLP:conf/kdd/ChangYZYD20,DBLP:journals/tkde/ZongS23}, and long-tailed distribution in MLTC~\cite{DBLP:conf/aaai/Xiao0JHS21,DBLP:conf/emnlp/HuangGKOO21,DBLP:journals/tois/YaoZZS24}.

\subsection{Nearest Neighbor based Retrieval-Augmented Method}
The retrieval augmented methods additionally use a datastore to retrieve a set of related documents to enhance the outputs of the model \cite{DBLP:journals/corr/abs-2202-01110,DBLP:conf/nips/LewisPPPKGKLYR020}.
The nearest neighbor method is a simple and effective retrieval-augmented method that uses the labels of the retrieved $k$ nearest neighbors to enhance the output.
Khandelwal et al.~\cite{DBLP:conf/iclr/KhandelwalLJZL20} introduce $k$NN-LM which extends a pre-trained neural language model by linearly interpolating it with the result of $k$NN.
The result of $k$NN is obtained by weighting the labels of $k$NN, where both nearest neighbors and weights are computed based on the similarity in the pre-trained embedding space.

Afterward, the nearest neighbor method is applied to various tasks, including machine translation~\cite{DBLP:conf/acl/ZhengZGHCLC20,DBLP:conf/naacl/YangS022,DBLP:conf/emnlp/JiangLMZZHS22,DBLP:conf/acl/WangFCX22}, MLTC~\cite{DBLP:conf/acl/SuWD22}, Chinese spelling check~\cite{DBLP:journals/corr/abs-2211-07843}, sequence labeling~\cite{DBLP:journals/corr/abs-2203-17103}, relation extraction~\cite{DBLP:conf/emnlp/WanLMCKL22}, and code vulnerability detection~\cite{DBLP:conf/emnlp/DuKZ22}.

\subsection{Contrastive Learning}
Contrastive learning is originally proposed in the computer vision community as a self-superivsed representation learning method~\cite{DBLP:journals/corr/abs-1807-03748OOrd,DBLP:conf/icml/ChenK0H20,DBLP:conf/cvpr/He0WXG20}.
Subsequently, some work use contrastive learning to improve the quality of sentence embeddings.
For example, ConSERT \cite{DBLP:conf/acl/YanLWZWX20} considers four data augmentation strategies and SimCSE \cite{DBLP:conf/emnlp/GaoYC21} uses dropout strategy to construct positive samples.
Khosla et al. \cite{khosla2020supervised} extend contrastive learning to supervised learning and use samples of the same class as positive.
Subsequently, contrastive learning has been used on a variety of tasks, including recommendation~\cite{DBLP:conf/sigir/YangHXL22}, entity set expansion~\cite{DBLP:conf/sigir/LiLHYS022}, collaborative filtering~\cite{DBLP:conf/sigir/LeePYL23}, and semantic concept embeddings~\cite{DBLP:conf/sigir/LiKBS23}.

Since some samples in MLTC do not have positive samples, there are challenges in how to use contrastive learning in MLTC.
Su et al. \cite{DBLP:conf/acl/SuWD22} treat samples with one same positive label as positive samples and do not require all labels to be identical.
They further design a dynamic coefficient based on the label similarity.
Lin et al. \cite{DBLP:conf/acl/LinQWZY23} further explore five different contrastive losses and found that strictly contrastive loss and Jaccard similarity contrastive loss perform better.
Jaccard similarity contrastive loss also treats samples with one same positive label as positive samples and uses Jaccard similarity to reweight the positive samples.
Strictly contrastive loss treats samples that are exactly the same as anchor labels as positive samples in the batch.
Besides, Wang et al.~\cite{DBLP:conf/acl/WangWH0W22} use label hierarchy to construct positive samples in hierarchical multi-label text classification.
Different from these methods, we use dropout strategy to construct clean positives and reweight negative samples based on label similarity to control the similarities of negatives.

\section{Methods}
In this section, we first present the task definition of MLTC. 
Then, we introduce the proposed DEbiased Nearest Neighbors (DENN) framework, followed by its technical details.
Finally, we conduct gradient analysis to illustrate the effectiveness of the weighted mechanism in contrastive learning.

\subsection{Problem Formulation}
Let $\mathcal{D}_{tr} = \{(\mathbf{x_i},\mathbf{y_i})\}_{i=1}^{N}$ be the MLTC training set consisting of $N$ samples, where $\mathbf{x_i}$ is raw text and $\mathbf{y_i} \in \{0, 1\}^C$ is corresponding label.
The task aims to learn a predictive model $f$: $\mathbf{x} \xrightarrow{} [0,1]^C$ that predicts scores close to 1 for positive labels and close to 0 for negative labels.
To facilitate the illustration, we list the main notations used throughout this article in Table~\ref{notation}.

\begin{table}[t]
\centering
\caption{The main notations used in the article, of which the upper part is used for the training phase and the lower part is used for the inference phase.}\label{notation}
\begin{tabular}{c|l}
\toprule
\textbf{Notation} & \textbf{Description}\\
\midrule
$\mathbf{h_i} \in \mathbb{R}^{d}$ &  Text representation for text $\mathbf{x_i}$\\
$d$ &  The dimension of feature representation\\
$\mathbf{W_1} \in \mathbb{R}^{C \times d}$ & Weight for classifier\\
$\mathbf{b_1} \in \mathbb{R}^{C}$ & Bias for classifier\\
$\mathbf{y_i} \in \mathbb{R}^{C}$ & Label of text $\mathbf{x_i}$  \\
$\mathbf{\hat{y}_i} \in \mathbb{R}^{C}$ & Predicted probability of classifier for text $\mathbf{x_i}$\\
$C$ &  The number of classes\\
$l_{ij} \in \mathbb{R}$ & label similarity between two texts $\mathbf{x_i}$ and $\mathbf{x_j}$\\
$w_{ij} \in \mathbb{R}$ & weight in contrastive learning for two texts $\mathbf{x_i}$ and $\mathbf{x_j}$\\
\midrule
$\mathbf{\hat{y}_{knn}} \in \mathbb{R}^{C}$ & Predicted probability of $k$NN\\
$\mathbf{\hat{y}_{clf}} \in \mathbb{R}^{C}$ & Predicted probability of classifier\\
$\mathbf{\bar{y}_{clf}} \in \mathbb{R}^{C}$ & Binary vectors indicate high-confidence prediction of $k$NN\\  
$\mathbf{\bar{y}_{knn}} \in \mathbb{R}^{M}$ & High-confidence prediction of classifier\\  
$\lambda \in \mathbb{R}$ & Debiased confidence of $k$NN\\
$M$ &  The number of high-confidence classes \\
\bottomrule
\end{tabular}
\end{table}

\subsection{Overview of Our Method}
\begin{figure*}[t]
\centering
\includegraphics[width=1\textwidth,height=0.6\textwidth]{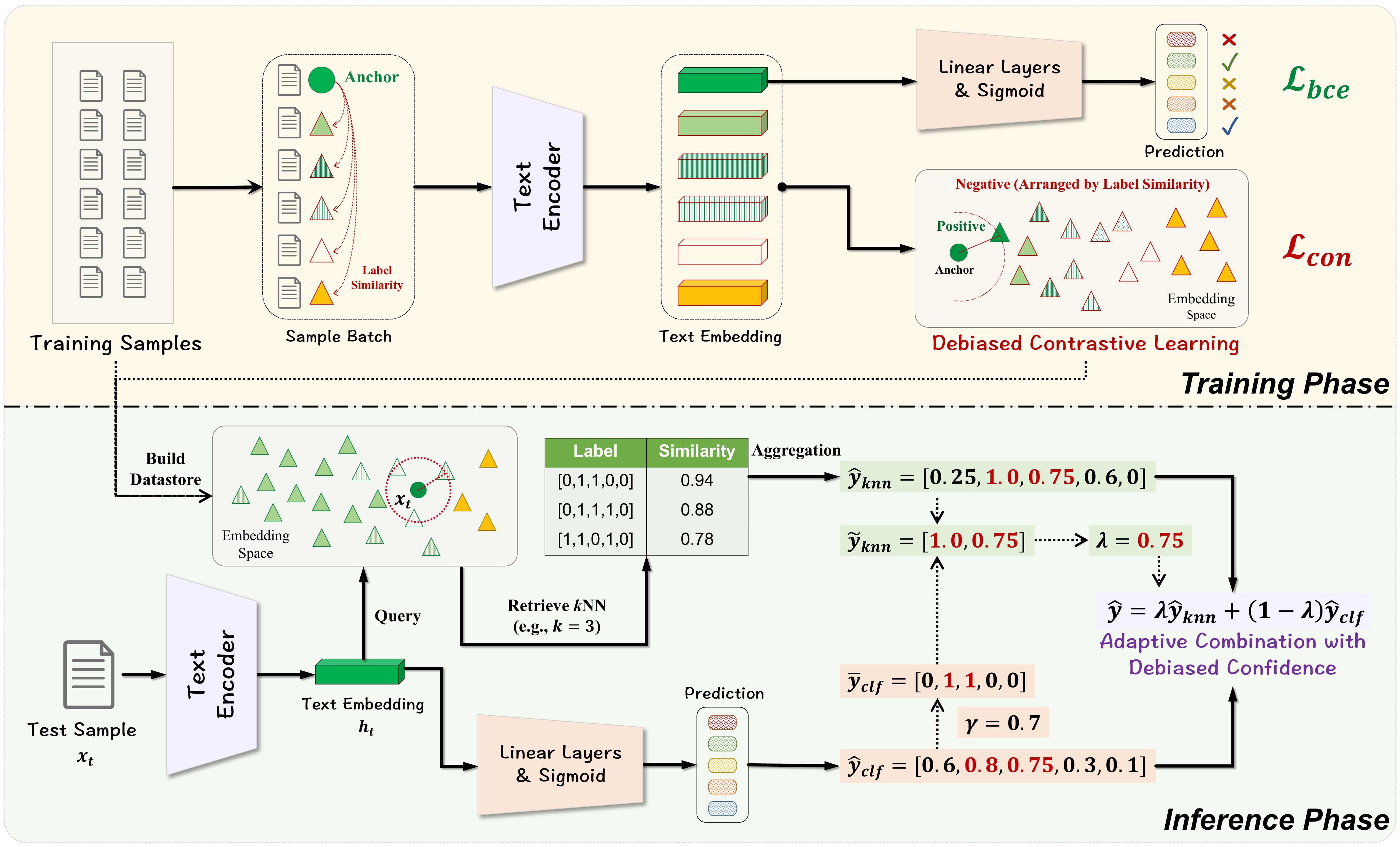}
\caption{Illustration of the Debiased Nearest Neighbors (DENN) Framework. The upper part denotes the training phase, where binary cross-entropy loss and debiased contrastive loss are used to train the model. The lower part represents the inference phase, which consists of three steps, building datastore, getting the two predictions (i.e., $\mathbf{\hat{y}_{knn}}$ and $\mathbf{\hat{y}_{clf}})$, and adaptive combining them with debiased confidence.} \label{fig:frame}
\end{figure*}

We propose a debiased nearest neighbors framework for MLTC as shown in Figure~\ref{fig:frame}.
As shown in the upper part of Figure \ref{fig:frame}, in the training phase, we use binary cross-entropy loss to train the text encoder and classifier and further propose debiased contrastive learning to adjust embedding space to maintain neighbor consistency for better $k$NN retrieve.
As shown in the lower part of Figure \ref{fig:frame}, in the inference phase, we first build datastore, get two outputs of the classifier and $k$NN, and further propose a debiased confidence estimation to adaptively combine two outputs.

\subsection{Text Encoder}
Given a text $\mathbf{x_i} = \{t_i^1, \cdots , t_i^n\}$ with $n$ tokens, we employ pre-trained encoder such as BERT \cite{DBLP:conf/naacl/DevlinCLT19} to extract the text representation:
\begin{equation}
\mathbf{h_i^0}, \mathbf{h_i^1}, ..., \mathbf{h_i^n}, \mathbf{h_i^{n+1}} = \textrm{Encoder}(\textrm{[CLS]}, t_i^1, ..., t_i^n, \textrm{[SEP]})
\end{equation}
where $\mathbf{h_i^0} \in \mathbb{R}^{d}$ is the representation of [CLS] token for text representation.
For simplicity, we use $\mathbf{h_i}$ to denote the text representation for text $\mathbf{x_i}$.

\subsection{Training with Debiased Contrastive Learning}
In MLTC, a model is usually trained by the binary cross-entropy (BCE) loss to classify text.
After using the encoder to extract text representation, we use a linear layer and sigmoid function to get probability distribution and then use BCE loss to train the encoder and classifier.
Specifically, the BCE loss for text $\mathbf{x_i}$ can be defined as follows:
\begin{equation}
    \mathbf{\hat{y}_i} = \text{sigmoid}(\mathbf{W_1}\mathbf{h_i}+\mathbf{b_1})
\end{equation}
\begin{equation}
    \mathcal{L}_{bce}^{(i)} = - \sum_{c=1}^C [y_i^c\text{log}\hat{y}_i^c + (1-y_i^c)\text{log}(1-\hat{y}_i^c)]
\end{equation}
where $\mathbf{\hat{y}_i} \in \mathbb{R}^{C}$ refers to the predicted scores of $\mathbf{x_i}$ on all labels, $\mathbf{W_1} \in \mathbb{R}^{C \times d}$ and $\mathbf{b_1} \in \mathbb{R}^{C}$ denote the weight and bias respectively.

However, the BCE loss is unaware of the $k$NN retrieval process.
As a result, the retrieved $k$NN may have lower label similarity with the test samples and provide little assistance for the prediction.
An ideal embedding space for $k$NN retrieval process should satisfy that the similarity of sample pairs is correlated with their label similarity.
The more similar the labels of the sample pairs are, the more similar their representations will be.
Inspired by contrastive learning can encourage embeddings from the same class to be pulled closer together and embeddings from different classes to be pushed apart~\cite{DBLP:conf/emnlp/KarpukhinOMLWEC20,DBLP:conf/acl/SuWD22}, we use contrastive learning to adjust embedding space to fit $k$NN retrieval.

There are two challenges in using contrastive learning for MLTC.
Firstly, since some samples in MLTC cannot find positive samples, we propose to construct clean positive samples using unsupervised data augmentation strategy such as dropout~\cite{DBLP:conf/emnlp/GaoYC21}.
This strategy ensures the quality of positives and avoids \emph{false positives} problem.
Secondly, vanilla contrastive learning simply pulls away all negatives without considering the label similarity between anchor and negatives.
Thus, some negatives with high embedding similarity will have a larger gradient to push away, even if they have high label similarity to the anchor~\cite{DBLP:conf/cvpr/WangL21a,DBLP:conf/emnlp/AnTCTZW22}.
Based on this, we further reweight negatives to correlate embedding similarity with label similarity.
A detailed gradient analysis can be found in Section \ref{sec:grad}.

Then, we propose debiased contrastive learning (DCL), which uses the dropout strategy to construct clean positives and reweights negatives based on label similarity.
During the training process of Transformers~\cite{DBLP:conf/nips/VaswaniSPUJGKP17}, each input will independently sample dropout masks placed on fully-connected layers and attention probabilities.
This means that even if the inputs are the same during training, the outputs will be different.
Thus, for positives, we simply feed the same input to the text encoder twice and get two embeddings (i.e., $\mathbf{h_i}$ and $\mathbf{h_i^+}$) with different dropout masks~\cite{DBLP:conf/emnlp/GaoYC21}.
Negatives with larger label similarity will have a smaller weight to avoid pushing them too far away and negatives with smaller label similarity will have a larger weight to push them away.
Specifically, we define the weight $w_{ij}$ in debiased contrastive learning by summing the weight in vanilla contrastive learning (i.e., 1) and negative label similarity (i.e., 1 - $l_{ij}$) as follows:
\begin{equation}
    w_{ij} = 1 + (1 - l_{ij})
\end{equation}
\begin{equation}
    l_{ij} =  \frac{\Vert \mathbf{y_i^\top}\mathbf{y_j} \Vert_{1}}{\text{max}(\Vert \mathbf{y_i} \Vert_{1}, \Vert \mathbf{y_j} \Vert_{1})}
\end{equation}
where $l_{ij}$ denotes the label similarity of the two samples calculated by dividing the number of common positive labels by the maximum number of positive labels, and $\Vert \Vert_{1}$ denotes $\ell_1$-norm.

We define the debiased contrastive learning for text $\mathbf{x_i}$ as follows:
\begin{equation}
    \mathcal{L}_{con}^{(i)} = - \text{log} \frac{ \text{exp}(s_{ii^+}/ \tau_1)}{\sum_{j=1}^{2N}\mathbbm{1}[j\ne i]w_{ij}\text{exp}(s_{ij}/ \tau_1)}
\end{equation}
where $s_{ii^+} = \text{sim}(\mathbf{h_{i}},\mathbf{h_{i^+}})$ denotes the cosine similarity of two representations, $\mathbf{h_{i^+}}$ denotes the text representation of the corresponding augmented positive of text $\mathbf{x_i}$, $\tau_1$ is the temperature, and $2N$ is the batch size after data augmentation.

Finally, we accumulate all samples in the batch to define the total loss function:
\begin{equation}
    \mathcal{L} = \sum_{i=1}^{2N} \mathcal{L}_{bce}^{(i)} + \alpha \mathcal{L}_{con}^{(i)}
\end{equation}
where $\alpha$ is a hyperparameter.

The whole training flow is illustrated in Algorithm~\ref{alg:train}.

\begin{algorithm}[t]
\caption{The Training Flow of Debiased Nearest Neighbors Framework} \label{alg:train}
\begin{algorithmic}
\Require The training set $\mathcal{D}_{tr}$
\Ensure A well-trained multi-label text classification model
\ForAll{iteration = 1, $\cdots$, MaxIter}
\State Randomly sample $N$ samples from $\mathcal{D}_{tr}$
\State $\rhd$ \textcolor{blue}{Base Encoder}
\State Feed $N$ samples into BERT twice to get the hidden states $\mathbf{h_1}, \cdots, \mathbf{h_{2N}}$
\ForAll{i = 1, $\cdots$, $2N$}
\State $\rhd$ \textcolor{blue}{Classification loss:}
\State Feed the hidden state into the linear layer to get the probability based on Eq. (2)
\State Calculate binary cross-entropy loss based on Eq. (3)
\State $\rhd$ \textcolor{blue}{Debiased contrastve loss:}
\State Compute the weight of contrastive loss based on Eq. (4) and Eq. (5)
\State Calculate contrastive loss based on Eq. (6)
\EndFor
\State $\rhd$ \textcolor{blue}{Optimization}
\State Obtain total loss based on Eq. (7) and update model
\EndFor
\State \Return The well-trained model
\end{algorithmic}
\end{algorithm}

\subsection{Inference with Debiased Confidence Estimation}
During the inference phase, we propose a debiased confidence estimation strategy that considers two predicted probability distributions to jointly estimate the confidence of $k$NN for adaptively combining two outputs.
The inference phase consists of three steps: using the training set to create datastore, getting two predictions of classifier and $k$NN, and adaptively combining two predictions with debiased confidence.

\textbf{Creating Datastore}\quad
The datastore is constructed offline using the training set and consists of a set of key-value pairs.
Specifically, given the $i$-th training data ($\mathbf{x_i}$, $\mathbf{y_i}$) $\in \mathcal{D}_{tr}$, we define the corresponding key-values pair ($\mathbf{k_i}$, $\mathbf{v_i}$), where $\mathbf{k_i}$ is the vector representation $\mathbf{h_i}$ extracted by text encoder and $\mathbf{v_i}$ is the label $\mathbf{y_i}$.
Specifically,

\begin{equation}
    (\mathcal{K}, \mathcal{V}) =  \{(\mathbf{h_i},\mathbf{y_i}) | (\mathbf{x_i}, \mathbf{y_i}) \in \mathcal{D}_{tr} \}
\end{equation}

\textbf{Getting Two Predictions}\quad
Given the test text $\mathbf{x_t}$, the model first generates a vector representation $\mathbf{h_t}$ and prediction of classifier $\mathbf{\hat{y}_{clf}}$ using text encoder and classifier.
Then, the text $\mathbf{x_t}$ queries the datastore with corresponding representation $\mathbf{h_t}$ to retrieve its $k$NN $\mathcal{N}$ according to cosine similarity.
Finally, we aggregate the labels of $k$NN to get the prediction of $k$NN.
Specifically, it computes a similarity distribution $\mathbf{\beta}$ over neighbors based on the softmax of their similarities and gets the $k$NN prediction $\mathbf{\hat{y}_{knn}}$ based on similarity distribution $\beta$ and labels of $k$NN.

\begin{equation}
\beta_i = \frac{e^{\text{sim}(\mathbf{h_t}, \mathbf{h_i})/ \tau_2}}{\sum_{j\in \mathcal{N}} e^{\text{sim}(\mathbf{h_t},\mathbf{h_j})/ \tau_2}}
\end{equation}
\begin{equation}
 \mathbf{\hat{y}_{knn}} = \sum_{i\in \mathcal{N}} \beta_i \mathbf{y_i}
\end{equation}
where $\text{sim}$(·,·) indicates cosine similarity, $\mathcal{N}$ is the set of retrieved $k$NN, and $\tau_2$ is the temperature.

\textbf{Adaptive Combination with Debiased Confidence}\quad
We propose a debiased confidence estimation strategy to estimate the confidence of $k$NN for adaptively combining two predictions. 
To quantify the confidence of $k$NN prediction, we evaluate its retrieved neighbors in terms of both label consistency and distribution concentration on a high-confidence label subset. 
Considering that the calculation of probabilities in $k$NN prediction already takes into account the information of labels and similarity, we can refer to such probabilities as the basis of confidence evaluation.
To reduce computational complexity and consider the relative confidence between two predictions, we choose to identify the high-confidence label subset from the classifier's prediction. 
Subsequently, we utilize the minimum probability of $k$NN prediction on this label subset as the holistic confidence of $k$NN prediction.
Empirically, the minimum probability yields better performance than the average probability.

Specifically, in the first step, we use predictions of the classifier and a threshold to determine the high-confidence label subset.
\begin{equation}
\bar{y}_{clf}^c = \mathbbm{1}[\hat{y}_{clf}^c \geq \gamma]
\end{equation}
where $\bar{y}_{clf}^c$ is a binary value indicating whether the label $c$ is high confidence and $\gamma$ is a hyper-parameter indicating the threshold.
A larger $\gamma$ often results in a smaller set of high-confidence label subset and usually yields a larger debiased confidence.

In the second step, we extract a corresponding high-confidence prediction vector $\mathbf{\tilde{y}_{knn}} = [\hat{y}_{knn}^{c_1}, \cdots$, $\hat{y}_{knn}^{c_M}$] from $\mathbf{\hat{y}_{knn}}$ based on whether the corresponding element in $\mathbf{\bar{y}_{clf}}$ is 1, where $c_1$ denotes the index of high-confidence class and $M$ denotes the number of high-confidence label.
Then, we use the minimum probability in $\mathbf{\tilde{y}_{knn}}$ to represent the debiased confidence $\lambda$ of $k$NN:
\begin{equation}
\lambda = \text{min}(\hat{y}_{knn}^{c_1}, \cdots, \hat{y}_{knn}^{c_M})
\end{equation}

Finally, the predictions of the classifier and $k$NN are combined with debiased confidence $\lambda$ of $k$NN to form the final prediction $\mathbf{\hat{y}}$.

\begin{equation}
\mathbf{\hat{y}} = \lambda \mathbf{\hat{y}_{knn}} + (1-\lambda) \mathbf{\hat{y}_{clf}}
\end{equation}

The whole inference flow is illustrated in Algorithm~\ref{alg:infer}.

\begin{algorithm}[t]
\caption{The Process of Inference with Debiased Confidence Estimation}\label{alg:infer}
\begin{algorithmic}
\Require The training set $\mathcal{D}_{tr}$, the test set $\mathcal{D}_{te}$, a well-trained model
\Ensure The final prediction $\mathbf{\hat{y}}$ for test set $\mathcal{D}_{te}$
\State $\rhd$ \textcolor{blue}{Creating Datastore:}
\State ($\mathcal{K}, \mathcal{V}$) = $\emptyset$
\ForEach{$(\mathbf{x_i},\mathbf{y_i})$ in $\mathcal{D}_{tr}$}
\State Feeding text $\mathbf{x_i}$ into text encoder to get text representation $\mathbf{h_i}$ based on Eq. (1)
\State Add $\{(\mathbf{h_i},\mathbf{y_i})\}$ 
\EndFor
\ForEach{$\mathbf{x_t}$ in $\mathcal{D}_{te}$}
\State $\rhd$ \textcolor{blue}{Getting Two Predictions:}
\State Feeding text $\mathbf{x_t}$ into text encoder to get text representation $\mathbf{h_t}$ based on Eq. (1)
\State Getting predictions of the classifier $\mathbf{\hat{y}_{clf}}$ by feeding text representation $\mathbf{h_t}$ into classifier based on Eq. (2)
\State Using text representation $\mathbf{h_t}$ to retrieve $k$NN $\mathcal{N}$ in the datastore according to cosine similarity
\State Getting predictions of $k$NN $\mathbf{\hat{y}_{knn}}$ by aggregating the labels of retrieved $k$NN based on Eq. (10)
\State $\rhd$ \textcolor{blue}{Adaptive Combination with Debiased Confidence:}
\State Using prediction of the classifier to estimate the high-confidence label subset $\mathbf{\bar{y}_{clf}}$ based on Eq. (11)
\State Extracting corresponding high-confidence prediction of $k$NN $\mathbf{\bar{y}_{knn}}$ based on $\mathbf{\hat{y}_{knn}}$ and $\mathbf{\bar{y}_{clf}}$
\State Using the minimum probability of high-confidence $k$NN prediction $\mathbf{\bar{y}_{knn}}$ to get debiased confidence $\lambda$ based on Eq. (12)
\State Using debiased confidence $\lambda$ to adaptively combine predictions of $k$NN $\mathbf{\hat{y}_{knn}}$ and classifier $\mathbf{\hat{y}_{clf}}$ to obtain final prediction $\mathbf{\hat{y}}$ based on Eq. (13)
\EndFor
\State \Return The final prediction $\mathbf{\hat{y}}$ for test set $\mathcal{D}_{te}$
\end{algorithmic}
\end{algorithm}

\subsection{Gradients Analysis}  \label{sec:grad}
In this section, we further analyze the gradients with respect to positive pairs and different negative pairs to show the effectiveness of our debiased contrastive learning.
Specifically, the gradients with respect to the positive similarity $s_{ii^+}$ and the negative similarity $s_{ij}$ are formulated as:
\begin{equation} \label{form:gradient}
\frac{\partial \mathcal{L}_{con}^{(i)}}{\partial s_{ii^+}} = 
- \frac{1}{\tau_1} \sum_{j=1}^{2N} \mathbbm{1}[j\ne i] \mathbbm{1}[j\ne i^+] P_{ij}
\end{equation}
\begin{equation}
\frac{\partial \mathcal{L}_{con}^{(i)}}{\partial s_{ij}} = \frac{1}{\tau_1} P_{ij}
\end{equation}
\begin{equation}
P_{ij} = \frac{ w_{ij}{\rm exp} (s_{ij}/\tau_1) }{ \sum_{k=1}^{2N} \mathbbm{1}[k\ne i] w_{ik}{\rm exp} (s_{ik}/\tau_1)}
\end{equation}

We can see that the gradient magnitude for positive is equal to the gradient sum of negatives and the gradient magnitude for negatives is proportional to the weight $w_{ij}$ and exponential similarity term $\text{exp}(s_{ij}/\tau_1)$~\cite{DBLP:conf/cvpr/WangL21a}.
However, all $w_{ij}$ are equal to 1 in vanilla contrastive learning.
Thus, the gradients for negatives in vanilla contrastive learning are only inversely affected by the similarity term and are independent of label similarity.
In contrast, our debiased contrastive learning arranges weights $w_{ij}$ based on label similarity $l_{ij}$ to achieve gradient magnitude inversely proportional to label similarity.

\section{Experiments}
In this section, we first introduce our experimental details, including datasets and evaluation metrics, experimental settings, and baselines. Then, we report the experimental results on four datasets to answer the following research questions:
\begin{itemize}
\item \textbf{RQ1}: Whether our proposed model outperforms existing multi-label text classification methods?
\item \textbf{RQ2}: How does each of the components of our model contribute to the final performance?
\item \textbf{RQ3}: Whether our proposed debiased contrastive loss outperforms other contrastive learning variants for multi-label text classification?
\item \textbf{RQ4}: How do the hyper-parameters and the size of datastore influence the performance of our method?
\item \textbf{RQ5}: Which classes are the main sources of performance improvement for our proposed model?
\end{itemize}
Thereafter, we visualize the learned embedding space and further analysis the time and space of our proposed approach.

\begin{table} \setlength{\tabcolsep}{8pt}
\caption{Statistics of the datasets. $\lvert$\textbf{$\mathcal{D}$}$\rvert$ and $\lvert$\textbf{$\mathcal{Y}$}$\rvert$ denote the number of samples and labels. $\lvert$\textbf{$\bar{\mathcal{Y}}$}$\rvert$ and $\lvert$\textbf{$\bar{\mathcal{W}}$}$\rvert$ denote the average number of labels and words per sample.}\label{statis}
\centering
\begin{tabular}{lcccccc}
\toprule
\textbf{Dataset}&$\lvert$\textbf{$\mathcal{D}$}$\rvert$&$\lvert$\textbf{$\mathcal{Y}$}$\rvert$&$\lvert$\textbf{$\bar{\mathcal{Y}}$}$\rvert$&$\lvert$\textbf{$\bar{\mathcal{W}}$}$\rvert$\\
\midrule
\textbf{AAPD} & 55840& 54& 2.41 &163.43 \\
\textbf{RCV1-V2} &  804414 &103&3.24&123.94 \\
\textbf{Amazon-531} & 49145 & 531& 2.93 & 98.43 \\
\textbf{EUR-LEX57K} &  57000 & 4271 & 5.07 & 757.96 \\
\bottomrule
\end{tabular}
\end{table}

\subsection{Datasets and Evaluation Metrics}
We validate our proposed model on four benchmark multi-label text classification datasets: AAPD~\cite{DBLP:conf/coling/YangSLMWW18}, RCV1-V2~\cite{DBLP:journals/jmlr/LewisYRL04}, Amazon-531~\cite{DBLP:conf/recsys/McAuleyL13}, and EUR-LEX57K~\cite{DBLP:conf/acl/ChalkidisFMA19}.
AAPD dataset contains the abstract and the corresponding subjects of 55,840 papers in the computer science field from the arXiv website.
RCV1-V2 dataset consists of over 800,000 manually categorized newswire stories made available by Reuters Ltd for research purposes.
Amazon-531 dataset contains 49,145 product reviews collected from Amazon and a three-level class taxonomy consisting of 531 classes.
It is worth noting that our paper focuses on multi-label text classification rather than hierarchy multi-label text classification and therefore does not use class hierarchy.
EUR-LEX57K contains 57k English EU legislative documents from the EUR-LEX portal, tagged with 4271 labels (concepts) from the European Vocabulary (EUROVOC).
We also splice header, recitals, main body to form the final text following ~\citet{DBLP:conf/acl/ChalkidisFMA19}.

For AAPD and RCV1-V2 datasets, we follow the widely adopted division by Yang et al.~\cite{DBLP:conf/coling/YangSLMWW18}.
For the Amazon-531 dataset, we follow the division by ~\citet{DBLP:journals/corr/abs-2204-03954}, i.e., we randomly sample 20\% of the training set to be the validation set.
For the EUR-LEX dataset, we follow the division by \citet{DBLP:conf/acl/ChalkidisFMA19}.
The statistics of these four datasets are listed in Table~\ref{statis}.

We use micro-F1 score and macro-F1 score as our evaluation metrics, micro-precision, micro-recall, macro-precision, and macro-recall are also reported for analysis.


\subsection{Experimental Settings}
We adopt base-uncased version of BERT \cite{DBLP:conf/naacl/DevlinCLT19} as the text encoder.
We use Adam optimizer~\cite{DBLP:journals/corr/KingmaB14}.
The batch size and learning rate are set to be 32 and 5e-5 on four datasets.
The maximum total input sequence length is 320 for AAPD, RCV1-V2, and Amazon-531 datasets.
The maximum total input sequence length is 520 for the EUR-LEX57K dataset.
$\alpha$ is set to 0.1 for the AAPD dataset, 0.3 for the RCV1-V2 dataset, 0.005 for the Amazon-531 dataset, and 0.001 for the EUR-LEX57K dataset.
$\tau_1$ and $\tau_2$ are both set to 0.05 for the AAPD, RCV1-V2 and EUR-LEX57K datasets.
$\tau_1$ is set to 0.02 and $\tau_2$ is set to 0.01 for the Amazon-531 dataset.
$\gamma$ is set to 0.7 for four datasets.
$k$ is set to 30 for the AAPD dataset, 40 for the RCV1-V2 dataset, 10 for the Amazon-531, and 20 for the EUR-LEX57K dataset.
We test the model with the best micro-F1 score on the validation set.

\begin{table*}[t]  \setlength{\tabcolsep}{3pt}
\caption{Comparison between our method and baselines on the AAPD and RCV1-V2 datasets. The best F1 score is \textbf{bold}. Avg. F1 denotes the average performance of micro-F1 and macro-F1 on two datasets. $\dag$ denotes that the results were reproduced by us because they did not use these datasets. * denotes the improvement on F1 is statistically significant (p $\textless$ 0.05) by comparing with \textbf{LACO}$_{plcp}$, \textbf{LACO}$_{clcp}$, and $k$\textbf{NN} in paired t-tests. ** denotes the improvement on F1 is statistically significant (p $\textless$ 0.05) by comparing with \textbf{CORE}, \textbf{LACO}$_{plcp}$, \textbf{LACO}$_{clcp}$, and $k$\textbf{NN} in paired t-tests.} \label{tab:performance_comparison}
  \centering
    \begin{tabular}{l|ccc|ccc|ccc|ccc|c}
    \toprule
    \multirow{3}{*}{\textbf{Method}}&\multicolumn{6}{c}{\textbf{AAPD}}&\multicolumn{6}{|c|}{\textbf{RCV1-V2}}&  \multirow{3}{*}{\tabincell{c}{\tabincell{c}{\textbf{Avg.}\\\textbf{F1}}}}\\
    \cmidrule(lr){2-7} \cmidrule(lr){8-13}
    & \multicolumn{3}{c|}{\textbf{Micro-}} &\multicolumn{3}{c|}{\textbf{Macro-}} &\multicolumn{3}{c|}{\textbf{Micro-}} &\multicolumn{3}{c|}{\textbf{Macro-}}\\
    & \textbf{P} & \textbf{R} & \textbf{F1} & \textbf{P} & \textbf{R} & \textbf{F1}& \textbf{P} & \textbf{R} & \textbf{F1}& \textbf{P} & \textbf{R} & \textbf{F1} \\
    \midrule
    \textbf{LSAN} & 77.7 & 64.6 & 70.6 & 67.6 & 47.2 & 53.5 & 91.3 & 84.1 & 87.5 & 74.9 & 65.0 & 68.4 & 70.0\\
    \textbf{BERT} & 78.6 & 68.7 & 73.4 & 68.7 & 52.1 & 57.2 & 92.7 & 83.2 & 87.7 & 77.3 & 61.9 & 66.7 & 71.3\\
    \textbf{CORE} & 80.3 & 70.4 & 75.0 & 70.4 & 54.6 & 59.5 &91.1 & 86.4 & \textbf{88.7} &75.9 & 68.4 & 70.3 & 73.4\\
    \midrule
    \textbf{Seq2Seq} & 69.8 & 68.2 & 69.0 & 56.2 & 53.7 & 54.0 & 88.5 & 87.4 & 87.9 & 69.8 & 65.5 & 66.1 & 69.3\\
    \textbf{SGM} & 74.8 & 67.5 & 71.0 & - & - & - & 89.7 & 86.0 & 87.8 & - & - & - & -\\
    \textbf{Seq2Set} & 73.9 & 67.4 & 70.5 & - & - & -  & 90.0 & 85.8 & 87.9 & - & - & - & -\\
    \textbf{OCD}  & - & - & 72.0 & - & - & 58.5  & - & - & - & - & - & - & -\\
     \textbf{SeqTag} & 74.3 & 71.5 & 72.9 & 61.5 & 57.5 & 58.5 & 90.6 & 84.9 & 87.7 & 73.7 & 66.7 & 68.7 & 72.0\\
    \textbf{ML-R} & 72.6 & 71.8 & 72.2 & - & - & -  & 89.0 & 85.2 & 87.1 & - & - & - & -\\
    \midrule
    $\textbf{LACO}_{plcp}$ & 79.5 & 70.8 & 74.9 & 68.4 & 55.8 & 59.9 & 90.8 & 86.2 & 88.4 & 76.1 & 66.5 & 69.2 & 73.1\\
    $\textbf{LACO}_{clcp}$ & 78.9 & 70.8 & 74.7 & 71.9 & 56.6 & 61.2  & 90.6 & 86.4 & 88.5 & 77.6 & 71.5 & 73.1 & 74.4\\
    $\textbf{SCL}^{\dag}$ & 74.9 & 73.2 & 74.0 & 62.3 & 58.5 & 58.9 & 88.1 & 87.1 & 87.6 & 72.8 & 68.8 & 69.0 & 72.4\\
    $\textbf{JSCL}^{\dag}$ & 76.0 & 72.0  & 73.9 & 63.7 & 56.6 & 58.1 & 87.4 & 87.0 & 87.2 & 72.1 & 71.9 & 70.5 & 72.4\\
    \midrule
    \textbf{$k$NN}  & - & - & 75.2  & - & - & - & - & - & 88.4 & - & - & - & -\\
    \midrule
\textbf{DENN (Ours)}& 77.5 & 75.2 &\bf{76.3}**& 67.6 & 61.4 & \textbf{62.2}** & 89.3  & 88.0 & \bf{88.7}* & 76.0& 74.1 & \bf{73.9}** & \bf{75.3}\\
    \bottomrule
    \end{tabular}
\end{table*}

\subsection{Baselines}
To evaluate the effectiveness of our method, we compare our method with four groups of baselines.
The first group of methods focuses on designing specific model structures to capture label correlations.
\begin{itemize}
\item \textbf{LSAN}~\cite{DBLP:conf/emnlp/XiaoHCJ19} adaptively learns the label-specific document representation with the aid of self-attention and label-attention mechanisms.
\item \textbf{BERT}~\cite{DBLP:conf/naacl/DevlinCLT19} directly uses BERT to encode text features and classify text.
\item \textbf{CORE}~\cite{DBLP:conf/ijcai/ZhangZYLCZ21} directly splices text and all labels and feeds them into BERT, and further uses the attention mechanism to highlight the most informative words in the document.
\end{itemize}

The second group of methods introduces special labeling schemes to capture label correlations.
\begin{itemize}
\item \textbf{Seq2Seq}~\cite{nam2017maximizing} proposes a sequence-to-sequence learning framework.
\item \textbf{SGM}~\cite{DBLP:conf/coling/YangSLMWW18} further adds attention mechanism and global embedding to the sequence-to-sequence learning framework.
\item \textbf{Seq2Set} \cite{DBLP:conf/acl/YangLMLS19} proposes a sequence-to-set model using reinforcement learning to train.
\item \textbf{OCD}~\cite{tsai2020order} uses the reinforcement learning algorithm based on optimal completion distillation to handle both exposure bias and label order in sequence generation.
\item \textbf{SeqTag}~\cite{DBLP:conf/acl/ZhangZYLC21} first obtains label embeddings based on label tokens in the \textbf{CORE} framework and then outputs a probability for each label sequentially by a BiLSTM-CRF model.
\item \textbf{ML-R}~\cite{DBLP:journals/ipm/WangRSQD21} proposes a novel iterative reasoning mechanism that takes account of the text and the label predictions from the previous reasoning step.
\end{itemize}

The third group of methods adds auxiliary tasks to capture label correlations.
\begin{itemize}
\item \textbf{{LACO}$_{plcp}$}~\cite{DBLP:conf/acl/ZhangZYLC21} uses an auxiliary task (i.e., predicts whether label pairs co-occur) based on \textbf{CORE} framework.
\item \textbf{{LACO}$_{clcp}$}~\cite{DBLP:conf/acl/ZhangZYLC21} predicts whether a label appears given a set of positive labels based on \textbf{CORE} framework.
\item \textbf{SCL} \cite{DBLP:conf/acl/LinQWZY23} uses a contrastive learning loss that treats samples that are exactly the same as anchor labels as positives in the batch.
\item \textbf{JSCL} \cite{DBLP:conf/acl/LinQWZY23} uses a contrastive learning loss treats samples with one same positive label as positives and uses Jaccard similarity to reweight the positives.
\end{itemize}

The fourth group of methods focuses on retrieving labeled training samples to capture label correlations.
\begin{itemize}
\item \textbf{$k$NN}~\cite{DBLP:conf/acl/SuWD22} uses BCE loss and dynamic contrastive loss with a coefficient to train the model. During the inference phase, the test document first retrieves $k$NN in the training set based on text representation and averages the predictions of the classifier and $k$NN.
\end{itemize}

\textbf{CORE}, \textbf{{LACO}$_{plcp}$}, and \textbf{{LACO}$_{clcp}$} need to concatenate text and labels to feed into BERT, thus they are unable to handle datasets with a large number of labels, i.e., Amazon-531 and EUR-LEX57K.
We search the hyper-parameters in $\textbf{SCL}$, $\textbf{JSCL}$, and $k\textbf{NN}$ for a fair comparison.
Specifically, the weight in contrastive learning is selected from \{0.0001, 0.0005, 0.001, 0.005, 0.01, 0.05, 0.1\} and the number of nearest neighbors is selected from \{5, 10, 20, 30, 40, 50\}.

\begin{table*}[t]  \setlength{\tabcolsep}{3.5pt}
\caption{Comparison between our method and baselines on the Amazon-531 and EUR-LEX57K datasets. \textbf{CORE}, \textbf{{LACO}$_{plcp}$}, and \textbf{{LACO}$_{clcp}$} need to concatenate text and labels to feed into BERT, thus they are unable to handle these two datasets with a large number of labels. The best F1 score is \textbf{bold}. Avg. F1 denotes the average performance of micro-F1 and macro-F1 on two datasets. $\dag$ denotes that the results were reproduced by us because they did not use these datasets. * denotes the improvement on F1 is statistically significant (p $\textless$ 0.05) by comparing with all baselines in paired t-tests.} \label{tab:performance_comparison2}
  \centering
    \begin{tabular}{l|ccc|ccc|ccc|ccc|c}
    \toprule
    \multirow{3}{*}{\textbf{Method}}&\multicolumn{6}{c|}{\textbf{Amazon-531}}&\multicolumn{6}{c|}{\textbf{EUR-LEX57K}} & \multirow{3}{*}{\tabincell{c}{\tabincell{c}{\textbf{Avg.}\\\textbf{F1}}}}\\
    \cmidrule(lr){2-7} \cmidrule(lr){8-13}
    & \multicolumn{3}{c|}{\textbf{Micro-}} &\multicolumn{3}{c|}{\textbf{Macro-}} &\multicolumn{3}{c|}{\textbf{Micro-}} &\multicolumn{3}{c|}{\textbf{Macro-}}\\
    & \textbf{P} & \textbf{R} & \textbf{F1} & \textbf{P} & \textbf{R} & \textbf{F1}& \textbf{P} & \textbf{R} & \textbf{F1}& \textbf{P} & \textbf{R} & \textbf{F1} \\
    \midrule
    \textbf{BERT$^{\dag}$} &90.6 & 88.4 & 89.5 & 64.5 & 57.3 & 59.1 & 78.6 & 72.1 & 75.2 & 30.7 & 26.8 & 27.6 & 62.9\\
    $\textbf{SCL}^{\dag}$ &  90.6 & 89.1 & 89.8 & 63.7 & 57.5 & 59.2 & 78.5 & 72.4 & 75.3 & 31.1 & 27.8 & 28.3 & 63.2\\
    $\textbf{JSCL}^{\dag}$ & 90.4 & 89.8 & 90.1 & 63.5 & 59.1 & 60.1 & 77.9 & 73.7 & 75.6 & 30.9 & 27.9 & 28.2 & 63.5\\
    \textbf{$k$NN$^{\dag}$}  &  90.5 & 90.0 & 90.2 & 63.2 & 62.8 & 61.9  & 78.3 & 72.5 & 75.3 & 30.8 & 27.7 & 28.1 & 63.9\\
    \midrule
\textbf{DENN (Ours)}& 90.6 & 90.6 & \bf{90.6}* & 65.5 & 63.9 & \bf{63.3}* & 78.3 & 73.7 & \bf{75.9}* & 31.5 & 28.6 & \bf{29.0}* & \bf{64.7}\\
    \bottomrule
    \end{tabular}
\end{table*}

\subsection{Results (RQ1)}
We conduct an empirical study to investigate whether \textbf{DENN} achieves better performance for MLTC on four datasets in Table \ref{tab:performance_comparison} and Table \ref{tab:performance_comparison2}.

Firstly, our proposed method achieves the best performance on the AAPD and RCV1-V2 datasets in Table \ref{tab:performance_comparison}.
Specifically, compared to strong baseline \textbf{$k$NN}, \textbf{DENN} achieves 1.1\% and 0.3\% micro-F1 improvement on two datasets.
Compared to strong baseline \textbf{LACO$_{clcp}$}, our model achieves 1.6\%, 1.0\%, 0.2\%, and 0.8\% improvement, respectively.
This shows that \textbf{DENN} is effective.
We further see that the improvement mainly comes from the recall metric.
This is because predictions of $k$NN recall more relevant labels.
In the three strong baseline (i.e., \textbf{$k$NN}, \textbf{CORE}, and \textbf{LACO$_{clcp}$}), \textbf{$k$NN} and \textbf{CORE} achieve the best micro-F1 on the AAPD and RCV1-V2 datasets, while \textbf{LACO$_{clcp}$} achieves the best macro-F1 on two datasets.
This shows that it is difficult for baselines to achieve the best results for both micro-F1 and macro-F1 on a dataset.
Our proposed \textbf{DENN} achieves the best performance on both two main metrics, indicating its effectiveness and consistent improvement.

Secondly, we compare our model with some strong baselines (i.e., \textbf{BERT}, \textbf{SCL}, \textbf{JSCL}, and \textbf{$k$NN}) on the Amazon-531 and EUR-LEX57K datasets in Table \ref{tab:performance_comparison2}.
It is worth noting that \textbf{CORE}, \textbf{{LACO}$_{plcp}$}, and \textbf{{LACO}$_{clcp}$} need to concatenate text and labels to feed into BERT, thus they are unable to handle these two datasets with a large number of labels.
This also shows that our method is more universally applicable, especially for datasets with a large number of labels.
Our proposed method also achieves the best performance on the Amazon-531 and EUR-LEX57K datasets.
Specifically, compared to strong baseline \textbf{$k$NN}, our model achieves 0.4\%, 1.4\%, 0.6\%, and 0.9\% improvement, respectively.
Compared to strong baseline \textbf{JSCL}, our model achieves 0.5\%, 3.2\%, 0.3\%, and 0.8\% improvement, respectively.
Our method also achieves improvement in both micro-F1 and macro-F1, indicating the effectiveness of \textbf{DENN}.
We found that the reason for the smaller improvement on the EUR-LEX57K dataset is limited by the performance of $k$NN.
Besides, the significant difference between micro-F1 and macro-F1 in these two datasets is due to the presence of a large number of few-shot and zero-shot labels.

On baselines, \textbf{LACO$_{clcp}$} further uses pre-trained task based on \textbf{CORE} and achieves better performance on recall metric in Table \ref{tab:performance_comparison}.
This shows that this conditional label co-occurrence prediction task can significantly improve recall but slightly decrease precision.
\textbf{SCL} and \textbf{JSCL} further use contrastive learning based on \textbf{BERT} and can be used for datasets with a large number of labels.
In most cases across the four datasets, both \textbf{SCL} and \textbf{JSCL} achieve performance improvements.
In addition, the improvement in macro-F1 is more significant compared to micro-F1.
This shows that contrastive learning is more effective for classes with fewer samples.
Due to the additional use of retrieved labels, \textbf{$k$NN} outperforms \textbf{SCL} and \textbf{JSCL} in most cases.

Finally, we compare the number of parameters of our model with some strong baselines in Table~\ref{tab:parameter}.
Since all models use the same backbone and classifier, we only compare the number of extra parameters.
We can see that our model achieves the best performance without extra parameters.
This shows our model is effective and efficient.
For baselines, using contrastive learning loss (i.e., \textbf{SCL} and \textbf{JSCL}) and $k$NN framework (i.e., $k$\textbf{NN}) does not introduce any extra parameters.
However, their performance improvement is relatively limited.
\textbf{CORE} uses extra convolutional layer to capture the relationship between words and labels.
\textbf{LACO$_{plcp}$} and \textbf{LACO$_{clcp}$} further use a linear layer to predict the relationship between two labels.
Although these baselines achieve better performance, they need extra parameters.

\begin{table}[t]
\caption{Extra number of parameters and performance of several strong baselines and DENN on two datasets.} \label{tab:parameter}
\begin{tabular}{l|cc|cc}
\toprule
\multirow{2}{*}{\textbf{Method}} & \multicolumn{2}{c|}{\textbf{AAPD}} & \multicolumn{2}{c}{\textbf{RCV1-V2}}  \\
         &  \textbf{Mi-F1/Ma-F1} & \textbf{\# Param}  & \textbf{Mi-F1/Ma-F1} & \textbf{\# Param} \\
         \midrule
\textbf{BERT}&  73.4/57.2  & \bf{0}  & 87.7/66.7 & \bf{0}  \\
\textbf{CORE}&  75.0/59.5  & 26298 & \textbf{88.7}/70.3 & 95584\\
\textbf{$\textbf{LACO}_{plcp}$}  & 74.9/59.9  & 27834 & 88.4/69.2 &  97120   \\
\textbf{$\textbf{LACO}_{clcp}$}  & 74.7/61.2 & 27834 & 88.5/73.1  &  97120   \\
\textbf{SCL}   & 74.0/58.9 & \bf{0}  & 87.6/69.0 & \bf{0}    \\
\textbf{JSCL}  & 73.9/58.1 & \bf{0}  & 87.2/70.5 & \bf{0}  \\
\textbf{$k$NN}   & 75.2/- & \bf{0}  & 88.4/- & \bf{0} \\
\textbf{DENN (Ours)}  & \bf{76.3}/\bf{62.2} & \bf{0}  & \bf{88.7}/\bf{73.9} & \bf{0}  \\ 
\bottomrule
\end{tabular}%
\end{table}

\subsection{Ablation Study (RQ2)}
In this section, we show the effectiveness of each component in our model by adding components one by one and report the performance of the classifier and $k$NN in Table~\ref{tab:abla}.

\begin{table*}[t]  \setlength{\tabcolsep}{5pt}
\caption{Ablation study on the two datasets. The best F1 score is \textbf{bold}. $\dag$ denotes the result is our implementation.}  \label{tab:abla}
  \centering
    \begin{tabular}{l|ccc|ccc|ccc|ccc}
    \toprule
    \multirow{3}{*}{\textbf{Method}}&\multicolumn{6}{c}{\textbf{AAPD}}&\multicolumn{6}{c}{\textbf{RCV1-V2}}\\
    \cmidrule(lr){2-7} \cmidrule(lr){8-13}
    & \multicolumn{3}{c}{\textbf{Mi- P / R / F1}} &\multicolumn{3}{|c|}{\textbf{Ma- P / R / F1}}& \multicolumn{3}{c}{\textbf{Mi- P / R / F1}} &\multicolumn{3}{|c}{\textbf{Ma- P / R / F1}}\\
    \midrule
    \multicolumn{13}{l}{\emph{Ablation study by adding components one by one:}}\\
    \textbf{BERT}$^{\dag}$& 76.1 & 71.6 & 73.8 & 65.5 & 55.6 & 57.1 & 88.2 & 86.6 & 87.4 & 72.0 & 70.0 & 69.6 \\
    \ \textbf{+$k$NN} & 77.2 & 72.8 & 74.9 & 64.5 & 57.5 & 58.9  & 89.1 & 86.4 & 87.7 & 73.2 & 69.2 &69.8\\
    \ \ \textbf{+DCL}& 77.9 & 74.3 & 76.1 & 68.2 & 59.1 & 61.1  & 90.1 & 87.0 & 88.5 & 76.3 & 71.3 &72.3\\
    \ \ \ \textbf{+debiased $\lambda$}& 77.5 & 75.2 &\bf{76.3}& 67.6 & 61.4 & \textbf{62.2} & 89.3  & 88.0 & \bf{88.7} & 76.0& 74.1 & \bf{73.9}\\
    \midrule
    \multicolumn{13}{l}{\emph{Ablation study on the performance of classifier and $k$NN:}}\\
    \textbf{Classifier}& 76.8 & 71.4 & 74.0 & 67.4 & 54.2 & 57.7 & 89.5 & 86.1 & 87.8 & 74.6 & 69.8 &70.8\\
    \textbf{$k$NN}& 76.9 & 74.3 & 75.6 & 65.9 & 61.3 & 61.4  & 89.4 & 87.2 & 88.3 & 74.7 & 72.1 & 72.0 \\
    \textbf{DENN}& 77.5 & 75.2 &\bf{76.3}& 67.6 & 61.4 & \textbf{62.2} & 89.3  & 88.0 & \bf{88.7} & 76.0 & 74.1 & \bf{73.9}\\
    \bottomrule
    \end{tabular}
\end{table*}

We observe the performance by adding components one by one.
Firstly, all components improve performance on two datasets.
This shows that all components are effective.
Secondly, we first add $k$NN retrieval during the inference phase, which achieves 1.1\% and 0.3\% micro-F1 improvement, and 1.8\% and 0.2\% macro-F1 improvement on two datasets.
This shows that combining the results of $k$NN can effectively improve performance.
Thirdly, we further add debiased contrastive learning, which achieves 1.2\% and 0.8\% micro-F1 improvement, and 2.2\% and 2.5\% macro-F1 improvement on two datasets.
This is because debiased contrastive learning improves the performance of the model and $k$NN by adjusting embedding space.
Finally, we add debiased $\lambda$, which further achieves 0.2\% and 0.2\% micro-F1 improvement, and 1.1\% and 1.6\% macro-F1 improvement on two datasets.
It is worth noting that the macro-F1 improvement is more significant.
This is because debiased $\lambda$ usually gives $k$NN bigger confidence and $k$NN performs better in macro-F1.

Then, we observe the performance of the classifier and $k$NN.
Firstly, DENN achieves the best performance.
This shows that combining two outputs is effective.
Secondly, the performance of $k$NN outperforms the classifier and the improvement comes mainly from the recall metric.
This shows that aggregating labels of $k$NN for prediction is more likely to predict more relevant labels.
In addition, the difference between the classifier and $k$NN in macro-F1 is significant compared to micro-F1.
This shows that $k$NN performs better on low-frequency classes compared to the classifier.
It is worth noting that \citet{DBLP:journals/ipm/CunhaMGCRNVFMAR21} has also found that non-neural methods can also achieve good performance.

\begin{table} \setlength{\tabcolsep}{12pt}
\caption{Performance of different contrastive losses on the AAPD dataset.}  \label{tab:contrastive}
\centering
\begin{tabular}{lccc}
\toprule
\textbf{Strategy}& \textbf{Mi-F1}&\textbf{Ma-F1} & \textbf{Average}\\
\midrule
\textbf{UCL} & 75.5 & 61.4 & 68.5\\
\textbf{SCL} & 74.0 & 59.5 & 66.8\\
\textbf{WSCL} & 75.0 & 60.1 &  67.6 \\
\textbf{Ours} & \bf{76.3} & \bf{62.2} & \bf{69.3} \\
\bottomrule
\end{tabular}
\end{table}

\subsection{Effects of Contrastive Learning (RQ3)}
In this section, we first explore the effects of various forms of contrastive learning loss and then investigate the effects of different positive sample generation techniques on the AAPD and RCV1-V2 datasets.

\subsubsection{Effects of Contrastive Loss}
We further explore three forms of contrastive loss.
The first is unsupervised contrastive learning (UCL), i.e., setting all $w_{ij}$ in Eq. (6) to 1.
The second is supervised contrastive learning (SCL), i.e., we first use the dropout strategy to ensure that each sample has a positive, and then use supervised contrastive loss \cite{khosla2020supervised} to train the model.
UCL treats augmented sample as positive, while SCL treats samples with the same label as positives.
The third one is weighted supervised contrastive learning (WSCL), i.e., we further reweight negatives based on SCL.

Firstly, as shown in Table \ref{tab:contrastive}, our debiased contrastive learning performs best.
This shows the effectiveness of debiased contrastive learning.
Secondly, unsupervised contrastive learning performs better than supervised contrastive learning.
This shows that it is difficult to learn to treat augmented sample and samples with identical labels equally as positives.
Secondly, weighting negatives improves performance for supervised and unsupervised contrastive learning.
This shows that using label similarity in contrastive learning can improve retrieval performance in two settings.

\begin{table} \setlength{\tabcolsep}{9pt}
\caption{Performance of different positive example generation techniques on the AAPD dataset.}  \label{tab:postive}
\centering
\begin{tabular}{lccc}
\toprule
\textbf{Strategy}&\textbf{Mi-F1}&\textbf{Ma-F1} & \bf{Average}\\
\midrule
\textbf{Random Masking} & 75.3 & 61.2 & 68.3 \\
\textbf{Continuous Masking}&   74.9 & 61.6 & 68.3\\
\textbf{Token Shuffling}&   73.6 & 57.7 & 65.7 \\
\textbf{Ours}&  \bf{76.3} & \bf{62.2} & \bf{69.3} \\
\bottomrule
\end{tabular}
\end{table}

\subsubsection{Effects of Positive Sample Generation}
We explore the effects of different positive sample generation techniques, including random masking, continuous masking, and token shuffling. 
Random/Continuous masking strategy randomly/continuously masks some of the tokens in the text.
Token shuffling strategy shuffles tokens in the text.

As shown in Table \ref{tab:postive}, our used dropout strategy achieves the best performance.
This shows that randomly discarding neuron-level information is the best way to construct effective positive samples.
Secondly, there is no significant performance difference between random masking and continuous masking.
Finally, token shuffling performs the worst.
This is because shuffling all tokens in the text is the most difficult strategy to recover the semantics of the text.

\begin{figure*}
  \centering
  \subfigure[The effects of $k$ on AAPD dataset.]{
        \includegraphics [width=0.235 \textwidth]{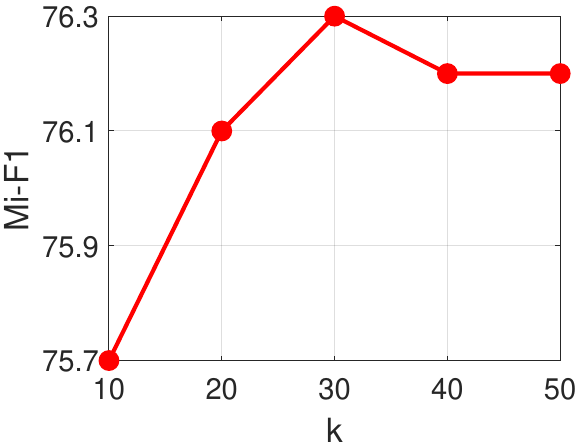}
 \includegraphics [width=0.235 \textwidth]{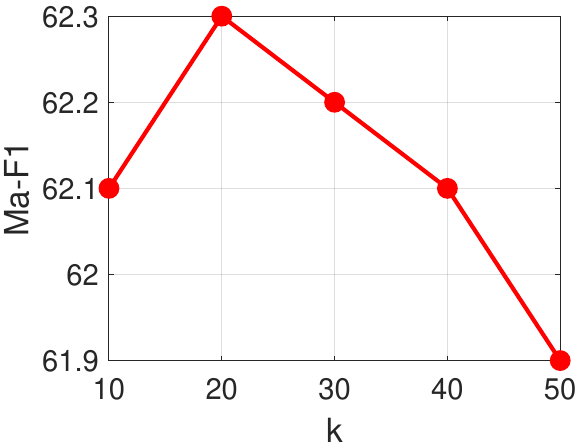}
  }
  \subfigure[The effects of $k$ on RCV1-V2 dataset.]{
         \includegraphics [width=0.235 \textwidth]{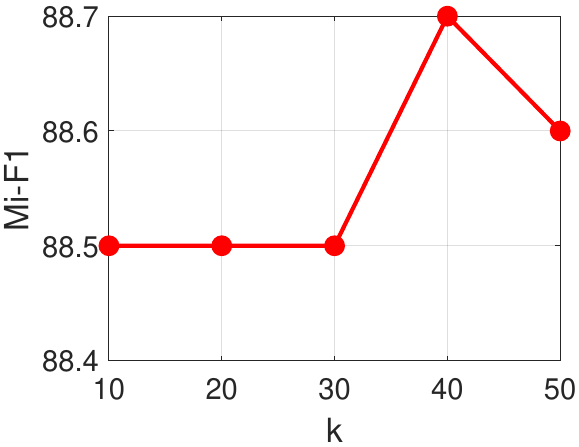}
  \includegraphics [width=0.235 \textwidth]{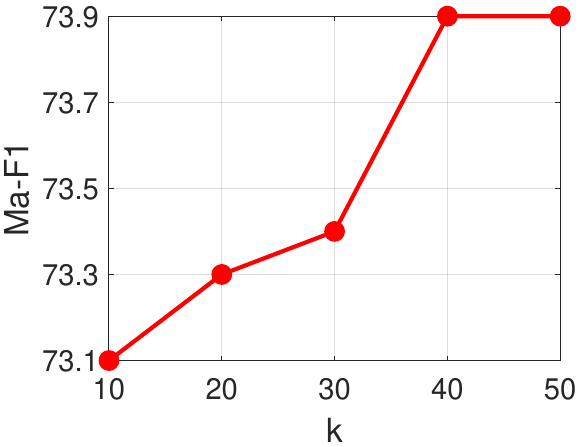}
  }
  \vspace{-1em}
  \caption{Effects of the number of retrieved nearest neighbors $k$ on two datasets.} \label{fig:k}
\end{figure*}

\subsection{Effects of Hyper-parameters and the Size of Datastore(RQ4)}
In this section, we explore the effects of hyper-parameters (i.e., the number of nearest neighbors $k$ and threshold $\gamma$) and the size of datastore on the AAPD and RCV1-V2 datasets.

\subsubsection{Effects of the Number of Nearest Neighbors}
We explore the effects of the number of retrieved nearest neighbors $k$ on two datasets in Figure \ref{fig:k}.

Firstly, the overall average performance of micro-F1 and macro-F1 increases first and then decreases, as $k$ increases.
This is because a small number of neighbors cannot fully utilize the knowledge in the datastore whereas a large number of neighbors leads to poor quality of neighbors.
When $k$ is 30, the average performance of micro-F1 and macro-F1 on the AAPD dataset is best.
When $k$ is 40, the average performance on the RCV1-V2 dataset is best.
This shows that the optimal number of nearest neighbors is different for different datasets.
The size of the datastore of the RCV1-V2 dataset is larger than AAPD, so the RCV1-V2 dataset needs a larger $k$.
Secondly, the overall performance fluctuates little as $k$ changes.
Specifically, micro-F1 and macro-F1 fluctuated by 0.6\% and 0.3\%, respectively, on the AAPD dataset.
Micro-F1 and macro-F1 fluctuated by 0.2\% and 0.8\%, respectively, on the RCV1-V2 dataset.
This shows that the performance is relatively insensitive as $k$ changes.

\begin{figure*}
  \centering
  \subfigure[The effects of $\gamma$ on AAPD dataset.]{
        \includegraphics [width=0.235 \textwidth]{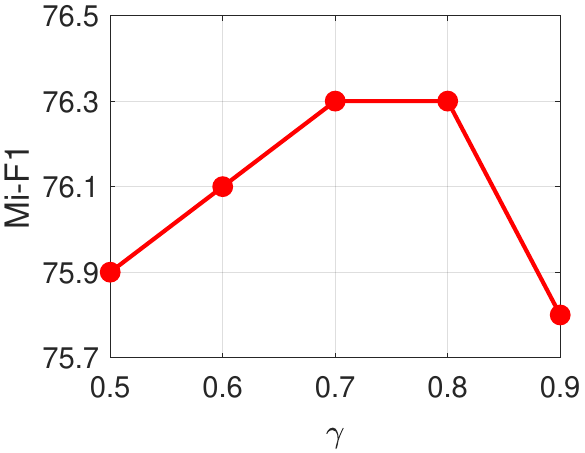}
 \includegraphics [width=0.235 \textwidth]{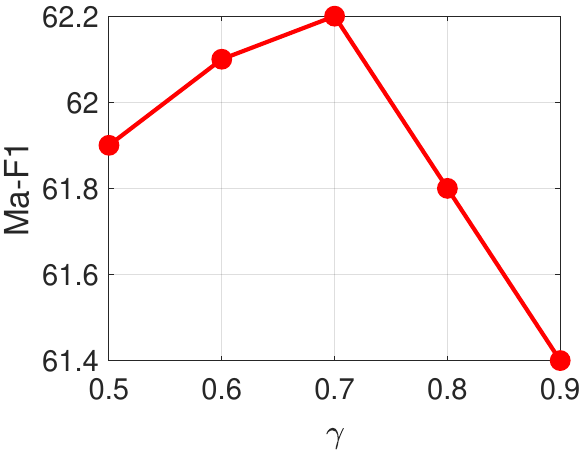}
  }
  \subfigure[The effects of $\gamma$ on RCV1-V2 dataset.]{
         \includegraphics [width=0.235 \textwidth]{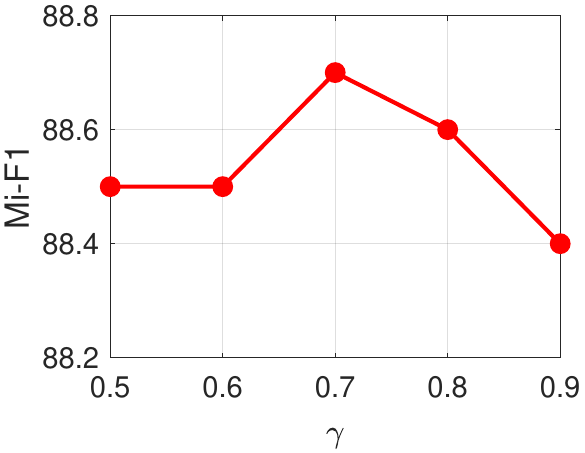}
  \includegraphics [width=0.235 \textwidth]{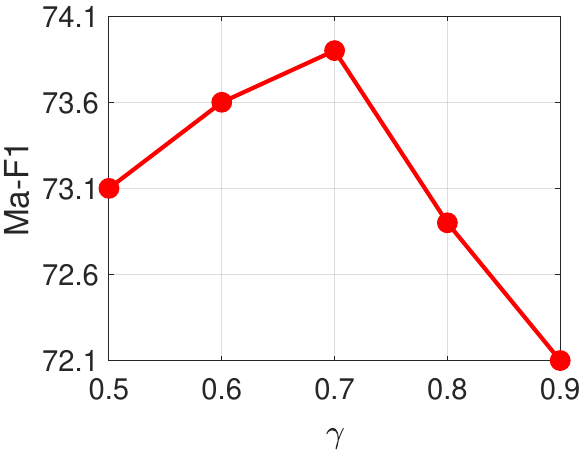}
  }
  \vspace{-1em}
  \caption{Effects of $\gamma$ on two datasets.} \label{fig:gamma}
\end{figure*}

\begin{figure*}
  \centering
  \subfigure[The effects of the size of datastore on AAPD dataset.]{
        \includegraphics [width=0.235 \textwidth]{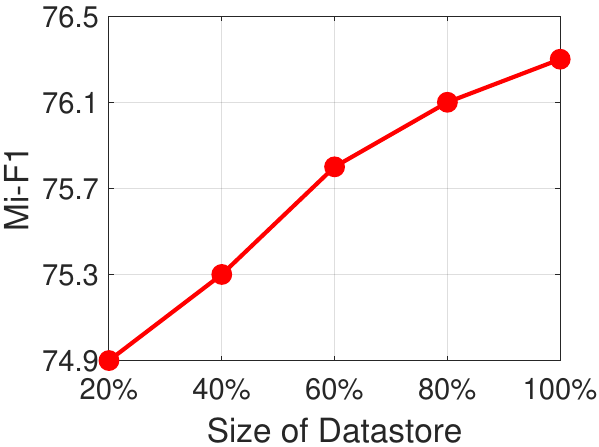}
 \includegraphics [width=0.235 \textwidth]{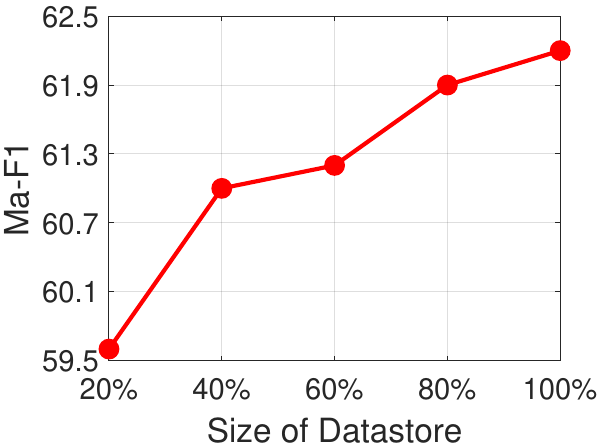}
  }
  \subfigure[The effects of the size of datastore on RCV1-V2 dataset.]{
         \includegraphics [width=0.235 \textwidth]{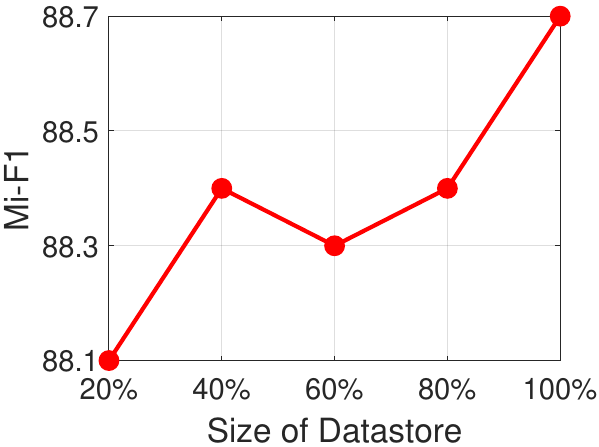}
  \includegraphics [width=0.235 \textwidth]{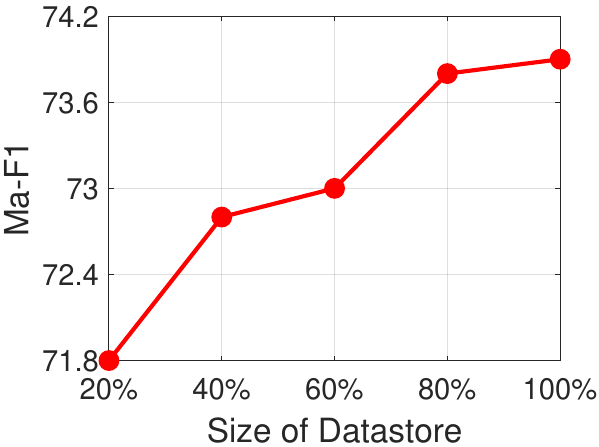}
  }
  \vspace{-1em}
  \caption{Effects of the size of datastore on two datasets.} \label{fig:datastore}
\end{figure*}

\subsubsection{Effects of $\gamma$}
We analyze the effects of the threshold $\gamma$ on two datasets in Figure \ref{fig:gamma}.

Firstly, the overall performance increases first and then decreases, as $\gamma$ increases.
When $\gamma$ equals 0.7, DENN achieves the best performance on two datasets.
We attribute this to two reasons:
Firstly, since the performance of $k$NN outperforms the classifier and the high-confidence labels estimated by a small $\gamma$ are unreliable, $\gamma$ needs to be greater than 0.5.
Secondly, when $\gamma$ is 0.8 or 0.9, the debiased confidence $\lambda$ will be larger, making it difficult to effectively combine the two outputs.

\subsubsection{Effects of the Size of Datastore}
We explore the effects of the size of the datastore on two datasets in Figure \ref{fig:datastore}.
We randomly sample a portion of the training set to construct a datastore to conduct the experiment, i.e., 20\%, 40\%, 60\%, 80\%, and 100\%.

Firstly, the overall performance gradually improves as the size of the datastore increases.
This is because larger repositories containing more knowledge are more helpful for performance improvement.
Secondly, the model achieves 1.4\% micro-F1 improvement and 2.6\% macro-F1 improvement on the AAPD dataset when the size of the datastore changes from 20\% to 100\%.
The model also achieves 0.6\% micro-F1 improvement and 2.1\% macro-F1 improvement on the RCV1-V2 dataset when the size changes from 20\% to 100\%.
We found a larger performance variation on the AAPD dataset.
This is because the size of the datastore of the RCV1-V2 dataset is larger, even though the ratio is 20\%.
Therefore, as the size of the datastore increases, the performance change is not significant on the RCV1-V2 dataset.

\begin{figure}[t]
\centering
\includegraphics[width=0.5\textwidth,height=0.3\textwidth]{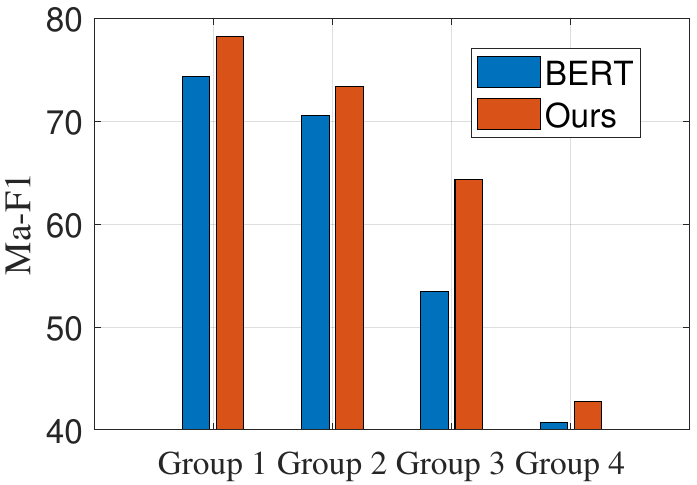}
\caption{Performance of BERT and DENN on four groups of classes with different frequencies on the AAPD dataset.} \label{fig:tail}
\end{figure}

\begin{figure}[t]
\centering
\includegraphics[width=0.6\columnwidth]{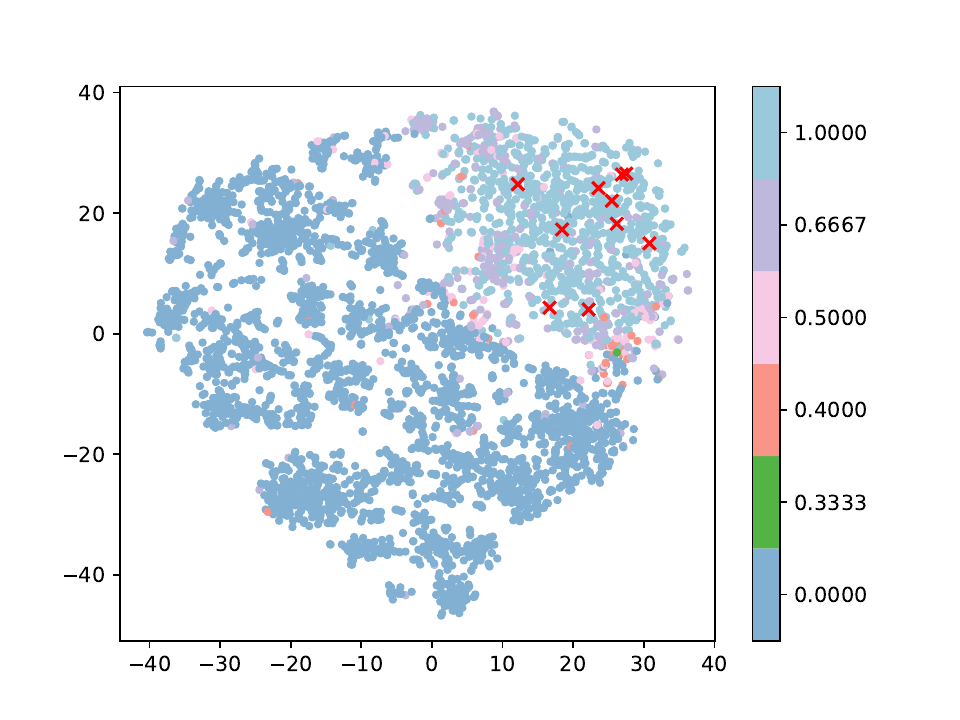}
\caption{Visualisation of the embedding space on the AAPD dataset. The red crosses indicate test samples with the same label, and other dots indicate training samples, where different colors indicate different label similarities to the test samples.}\label{fig:embedding}
\end{figure}

\subsection{Performance on Classes with Different Frequencies (RQ5)}

In this section, we compare our method with BERT on four classes to explore which classes are the main sources of performance improvement for our proposed model on the AAPD dataset.
We divide all the labels into four groups according to frequency, i.e., Group 1 (frequency$\textgreater$4500), Group 2 (4500$\geq$frequency$\textgreater$1700), Group 3 (1700$\geq$frequency$\textgreater$870), and Group 4 (870$\geq$frequency).

Firstly, as shown in Figure \ref{fig:tail}, our model achieves significant improvements in all four groups.
This shows that the improvement in classes is general and independent of frequency.
Specifically, our method achieves 3.9\%, 2.8\%, 10.9\%, and 2\% improvement, respectively.
Secondly, our model achieves the maximum performance gains in Group 3.
This indicates that for these classes with a certain sample size but insufficient, the improvement is most significant.
Our model also achieves improvement in Group 4 that are difficult to learn, indicating that our method is also effective in classifying low-frequency classes.

\subsection{Visualization}
In this section, we visualize training samples and 10 test samples with the same label to show the distribution of training samples and test samples on the AAPD dataset.

Firstly, the test samples have the same label (i.e., label similarity is 1) as the neighbors.
This shows that utilizing the labels of $k$NN is effective.
Secondly, training samples with high label similarity to the test samples tend to be closer to the test samples.
Overall, the colors of the training samples closer to the test samples are light blue, purple, pink, orange, green, and deep blue.
This shows that the distance from negative samples to test samples is related to label similarity.
This demonstrates the effectiveness of weighted mechanism in our proposed debiased contrastive learning.

\subsection{Analysis of Time and Space Overhead}
In this section, we analyze the time and space overhead introduced by $k$NN retrieval in our model.

As shown in Table \ref{tab:space}, the extra inference time per text increased by 12\% compared to w/o $k$NN and does not exceed 0.4 ms per text.
This shows that $k$NN retrieval is fast.
Secondly, the size of the datastore on the AAPD dataset is 0.165 GB.
This shows that the overall space overhead is acceptable.
Overall, $k$NN retrieval achieves performance improvement, and the time and space overheads are acceptable.


\begin{table} \setlength{\tabcolsep}{8pt}
\caption{Time and space overhead of two methods on the AAPD dataset. Inference time (ms/text) is tested with four Tesla V100 GPUs.}  \label{tab:space}
\centering
\begin{tabular}{lcccccc}
\toprule
\textbf{Method}&\textbf{Inference Time} &\textbf{Space}\\
\midrule
\textbf{w/o $k$NN}& 2.45 ms & 0 GB\\
\textbf{Ours}& 2.76 ms & 0.165 GB\\
\bottomrule
\end{tabular}
\end{table}

\section{Conclusion and Future Work}
In this paper, we propose a debiased nearest neighbors framework for MLTC.
To solve the embedding alignment bias and confidence estimation bias in the $k$NN framework, we propose a debiased nearest neighbors framework, including a debiased contrastive learning strategy and a debiased confidence estimation strategy.
The debiased contrastive learning strategy aims to enhance neighbor consistency and the debiased confidence estimation strategy aims to adaptively combine two outputs.
Extensive experiments show the effectiveness and efficiency of our proposed method.

In the future, we plan to deepen and widen our work from the following aspects:
(1) In this work, we propose a debiased contrastive learning to avoid false positives and control the distribution of negatives.
However, it is worth exploring whether there exist other more effective contrastive learning strategies for multi-label text classification.
(2) Our proposed contrastive learning uses the unsupervised dropout strategy to construct positives.
It is worth exploring how to design positive samples for multi-label text classification.
(3) The effectiveness of $k$NN framework on text classification also deserves further exploration.

\bibliographystyle{ACM-Reference-Format}
\bibliography{sample-base}

\appendix

\end{document}